# Facial Emotion Recognition Under Mask Coverage Using a Data Augmentation Technique

Aref Farhadipour
Media Engineering
IRI Broadcasting University
Tehran, Iran
areffarhadi@gmail.com

Pouya Taghipour
Microelectronics and Communications Research Lab
École de technologie supérieure (ETS)
Montréal, Quebec, Canada
pouya.taghipour.1@ens.etsmtl.ca

*Abstract*— **Identifying human emotions using AI-based computer vision systems, when individuals wear face masks, presents a new challenge in the current Covid-19 pandemic. In this study, we propose a facial emotion recognition system capable of recognizing emotions from individuals wearing different face masks. A novel data augmentation technique was utilized to improve the performance of our model using four mask types for each face image. We evaluated the effectiveness of four convolutional neural networks, Alexnet, Squeezenet, Resnet50 and VGGFace2 that were trained using transfer learning. The experimental findings revealed that our model works effectively in multi-mask mode compared to single-mask mode. The VGGFace2 network achieved the highest accuracy rate, with 97.82% for the person-dependent mode and 74.21% for the person-independent mode using the JAFFE dataset. However, we evaluated our proposed model using the UIBVFED dataset. The Resnet50 has demonstrated superior performance, with accuracies of 73.68% for the person-dependent mode and 59.57% for the person-independent mode. Moreover, we employed metrics such as precision, sensitivity, specificity, AUC, F1 score, and confusion matrix to measure our system's efficiency in detail. Additionally, the LIME algorithm was used to visualize CNN's decision-making strategy.**

**Keywords— Facial Emotion Recognition; Convolutional Neural Network; Data Augmentation; Transfer Learning**

## I. INTRODUCTION

The popularity of computer assistant systems is dependent on the proper interaction between humans and machines, regardless of real-world scenarios. Emotion recognition from facial expressions plays a crucial role in ensuring accurate communication between humans and machines. However, expressing emotions through the face involves variations in the form and position of facial organs. Additionally, real-world obstacles, such as a mask covering the face, may hinder the system from correctly interpreting a person's emotional state entirely.

The COVID-19 pandemic has resulted in the widespread use of masks in crowded environments, such as workplaces and hospitals. Despite the increased use of masks, individuals still wish to avail of services that rely on machine interaction. Unfortunately, the standard emotion recognition systems are unable to correctly monitor the person's entire face due to partial mask coverage.

To address this issue, we propose a new emotion recognition system that can perform effectively in the presence of masks. We first trained the system using datasets comprising images of individuals wearing masks. Convolutional Neural Networks (CNN) were used in four different architectures, and applied with the transfer learning approach. To prevent overfitting and address data scarcity challenges [1], we introduce an augmentation technique that adds different masks to the face to increase image variations. This approach involves using an automated algorithm to place mask models on the human face.

To the best of our understanding, only a few studies have been done on facial emotion recognition under masked conditions. Castellano et al. [2] focused on emotion recognition task solely from the eye area. This approach was assessed using the masked FER-2013 dataset containing seven emotions. In [3], a deep network with a two-stage attention mechanism was proposed to tackle the obstacles of face masks in emotion recognition. The network could recognize three emotions: positive, negative, and neutral.

Paper [4] proposed face expression recognition system based on vision transformers. Based on this strategy in the first step, a face parsing model has been trained to recognize better covered part of the face from the naked. In the next step, they established a vision transformer extractor for face emotion recognition and in the best situation, 62% accuracy was reported on the masked FER-2023 dataset.

In [5], a variety of training schemes have been evaluated to better understand the changes in arusal and valence dusing the masked facial emotion expression. The best results on the masked AffectNet dataset were obtained, with 53% and 45% accuracy for arousal and valence, respectively. Magherini et al. [6] utilized Resnet50 and InceptionResnet to classify five-class emotion recognition for masked images. Their system achieved 96.92% accuracy on AffectNet dataset. Dinca et al. [7] proposed an autoencoder artichecture of CNN to recognize positive and negative emotions, yielding 95.4% for the AFEW-VA dataset.

## II. EXPERIMENTAL SETUP

Datasets are a crucial requirement in the design of pattern recognition systems. In this paper, we used two datasets, namely JAFFE and UIBVFED. We will introduce these datasets, their preprocessing algorithms, and the convolutional neural networks utilized in this work.

## A. UIBVFED Dataset

The dataset used in this study comprises avatar images, featuring 20 virtual characters of 10 men and 10 women of varying skin colors and age groups ranging from 20 to 80 years old. The dataset only includes one type of mask and unbalanced images of the seven basic emotions [8]. Sample images of the UIBVFED are depicted in Fig. 1.

The dataset provides only 20 images for neutral and surprise, whereas happiness has 280 images and sadness has 120 images. The asymmetry in the dataset is due to some of the basic emotion categories consisting of compound emotions. As a result, two emotions, neutral and surprise, were removed to train the networks in a balanced manner. For the emotion of happiness, four compound emotions, namely "AbashedSmile", "DebauchedSmile", "EagerSmile", and "SlySmile", were removed. Similarly, for the sadness category, two compound emotions, "CryingClosedMouth" and "Miserable" were removed. To summarize, the dataset includes 80 images for each of the five basic emotions, including anger, disgust, fear, happiness and sadness.

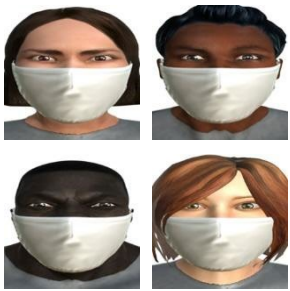

Fig. 1. Image samples of the UIBVFED dataset

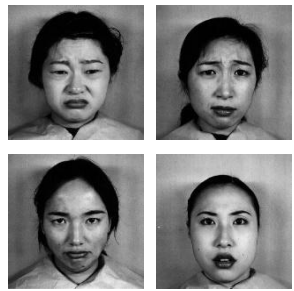

Fig. 2. Image samples of the JAFFE dataset

## B. JAFFE Dataset

The JAFFE dataset contains 213 images of 10 different Japanese female subjects. Each subject presents seven facial emotions, including neutral, fear, happiness, anger, disgust, sadness, and surprise. Each person expressed each emotion almost three times, and the images were annotated by 60 annotators [9]. Fig. 2 presents sample images from this dataset. Note that the JAFFE dataset contains unmasked facial emotions, which require occlusion by the mask.

## C. Proposed Data Augmentation

Fig. 3 depicts the CNN-based face emotion recognition under the mask coverage. Our proposed data augmentation method was applied to the JAFFE dataset. This method generates new training data using different types of masks to cover the face. In this situation, we have a wide variety of images to train the classifier. We inspired this method by adding noise and music to audio signals in speech processing scenarios as an augmentation method. As the JAFFE dataset includes backgrounds that are not required, and doesn't come with a mask, it's necessary to mask the face and crop the image to only the face part before applying classification techniques. We used the MaskTheFace algorithm [10] to artificially cover the face with four types of masks: surgical, cloth, N95, and KN95. Once the mask was applied, the MTCNN algorithm [11], a deep learning-based method for detecting faces within images, was employed to remove the unwanted, non-face background regions. Both MaskTheFace and MTCNN are well-known for their effectiveness in deep learning-based approaches for computer vision tasks.

## D. Transfer Learning

In transfer learning, a pre-trained network is adapted to a new scenario by adjusting the network's weights. CNNs typically consist of two parts: feature extraction and classification. In transfer learning, the last layers of the network that perform the classification are replaced with new layers to learn the features of the new classes. In this paper, we utilize four pre-trained CNNs: Alexnet [12], Squeezenet [13], Resnet50 [14], and VGGFace2 [15]. Alexnet consisted of 25 layers and was trained on 1000 classes from the Imagenet dataset [16]. Squeezenet is inspired by the Network in Network (NiN) architecture [17] and was also trained on the Imagenet dataset for 1000 classes. Resnet50 contains 50 layers of depth, about 25M learnable parameters, and proposed residual connections. VGGFace2 included 180 layers in the Resnet50 architecture and was trained on 3.31 million images from the VGGFace2 dataset for face recognition. We employ this pre-trained CNN for its strong background knowledge of facial features.

## E. Evaluation Metrics

There are various metrics available to evaluate a system, and selecting appropriate parameters would showcase the system's efficiency from different perspectives and allow for comparison with other works. The proposed evaluation parameters include precision, specificity, sensitivity, accuracy, and F1 score. These parameters are based on four indicators: True Positive (TP), which indicates correctly accepted, True Negative (TN), which expresses correctly rejected, False Positive (FP), which confirms incorrectly accepted, and False Negative (FN), which indicates incorrectly rejected. All four indicators are calculated based on a decision threshold value that could influence the results. These parameters can be expressed in the confusion matrix. The metrics for these indicators can also be formulated accordingly [18].

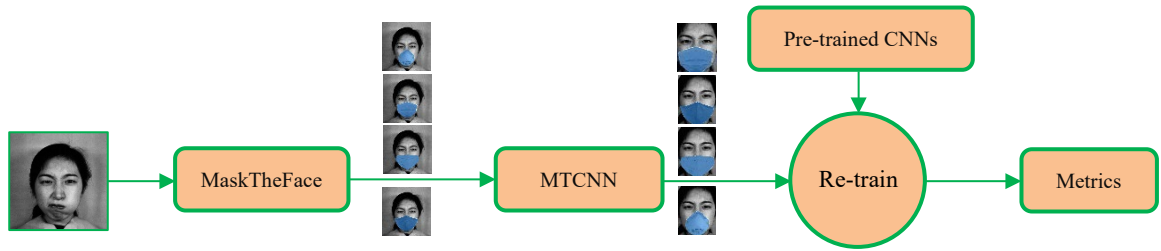

Fig. 3. Blockdiagram of CNN based system with proposed data augmentation

$$Specificity = \frac{TN}{TN+TP}$$

$$Precision = \frac{TP}{TP+FP}$$

$$Sensitivity = \frac{TP}{TP+FN}$$

$$Accuracy = \frac{TP+TN}{TP+TN+FP+FN}$$

$$F1\ Score = \frac{TP}{TP+\frac{1}{2}(FP+FN)}$$

In addition to these parameters, the ROC curve is drawn, and the AUC is also calculated for all proposed systems. Changing the value of the decision threshold makes the ROC diagram, and the AUC expresses the area under the ROC curve. Moreover, the confusion matrix is reported for each system.

## III. EXPERIMENTAL RESULT

This study conducted evaluations using two methods: Person-Dependent (PD) and Person-Independent (PI). In the PD method, the identities of the individuals in the testing and training phases were the same, while in the PI method, they were different. Two datasets were used for the investigations, namely the JAFFE and UIBVFED datasets.

For the JAFFE dataset, three individuals with IDs KM, NM, and YM were selected for testing in the PI mode. In the PD section, the third session of each person was considered as the test file, while the other two performances were used for training.

In the case of the UIBVFED dataset, five identities - Alicia, Jose, Ramon, Tomeu, and Wanda - were used for testing in the PI mode, while the remaining identities were used for training. In the PD mode, 75% of the data was used for training and 25% for testing.

In addition to our proposed data generation method to manipulate the training data with adding different mask to the face, two common data augmentation techniques were used: rotation operators within the range of [-20, 20] degrees, and vertical and horizontal translation by a distance of [5, 5] pixels. Moreover, we employed four types of masks for the JAFFE dataset, resulting in a total of 213x4 images. If an original image with one type of mask was used for training, that same image with the other three masks would not be included in the test set. Notably, the learning rate was set to 0.001 for all experiments, and the PD mode had an epoch number of 25 while the PI mode had 60.

In the first experiment, we trained Alexnet using only the surgery mask to evaluate the system efficiency without the proposed augmentation method. This evaluation was repeated for both PD and PI modes, and the results are presented in Table I. Confusion matrices for both modes are shown in Fig. 4, providing a detailed view of the network's performance for each class.

TABLE I. RESULT OF ALEXNET IN THE CASE OF USING A SINGLE TYPE OF MASK IN PD AND PI MODES

| Mode | Accuracy | Sensitivity | Specificity | Precision | F1 Score | AUC |
|---|---|---|---|---|---|---|
| *PD* | 85.507 | 0.855 | 0.976 | 0.855 | 0.855 | 0.968 |
| *PI* | 60.938 | 0.609 | 0.935 | 0.609 | 0.609 | 0.811 |

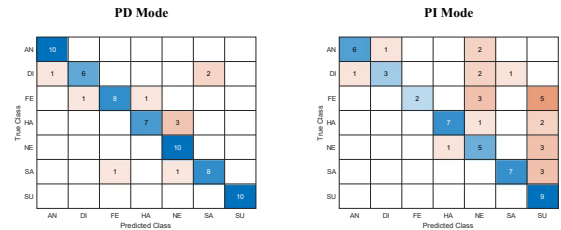

Fig. 4 Confusion matrixes for single-mask in PD and PI modes

The system trained in PD mode achieves an accuracy of 85.8%, while the PI mode acquires an accuracy of 60.9%. The confusion matrix reveals that the SU and NE classes are responsible for most of the errors in the PI mode. Due to the reduced identity changes in test and training images in PD mode, the system performs better in this mode than in PI mode.

In the second experiment, we evaluated three networks - Alexnet, Squeezenet, and VGGFace2 - using four types of masks in both PD and PI modes. Tables (II) and (III) present the results of the proposed systems on the JAFFE dataset in PD and PI modes, respectively. We also showed the confusion matrices for all six trained networks in Fig. 5.

TABLE II. RESULTS OF MULTI-MASK SCENARIO IN PD MODE

| Pre-trained Net | Accuracy | Sensitivity | Specificity | Precision | F1 Score | AUC |
|---|---|---|---|---|---|---|
| *AlexNet* | 93.543 | 0.935 | 0.989 | 0.941 | 0.933 | 0.997 |
| *SqueezeNet* | 92.391 | 0.924 | 0.987 | 0.924 | 0.924 | 0.994 |
| *ResNet50* | 93.841 | 0.938 | 0.990 | 0.938 | 0.938 | 0.938 |
| *VGGFace2* | 97.826 | 0.978 | 0.996 | 0.978 | 0.978 | 0.999 |

TABLE III. RESULTS OF MULTI-MASK SCENARIO IN PI MODE

| Pre-trained Net | Accuracy | Sensitivity | Specificity | Precision | F1 Score | AUC |
|---|---|---|---|---|---|---|
| *AlexNet* | 69.922 | 0.699 | 0.950 | 0.699 | 0.699 | 0.877 |
| *SqueezeNet* | 67.188 | 0.672 | 0.945 | 0.672 | 0.672 | 0.890 |
| *ResNet50* | 70.703 | 0.707 | 0.951 | 0.707 | 0.707 | 0.852 |
| *VGGFace2* | 74.219 | 0.742 | 0.957 | 0.742 | 0.742 | 0.869 |

Based on the multi-mask mode results, the VGGFace2 network achieved the highest performance in PD mode with a 97.82% accuracy rate. The sensitivity, specificity, precision, F1 score, and AUC were 0.978, 0.996, 0.978, 0.978, and 0.999, respectively. However, in PI mode, these evaluation parameters experienced a significant drop, resulting in an accuracy of 74.21%. Further examination of the confusion matrices in PI mode enables detailed performance assessment for each network. In comparison with one of the recent works [6], which reported 96.92% accuracy, our proposed system in the best situation achieved 0.9% outperform.

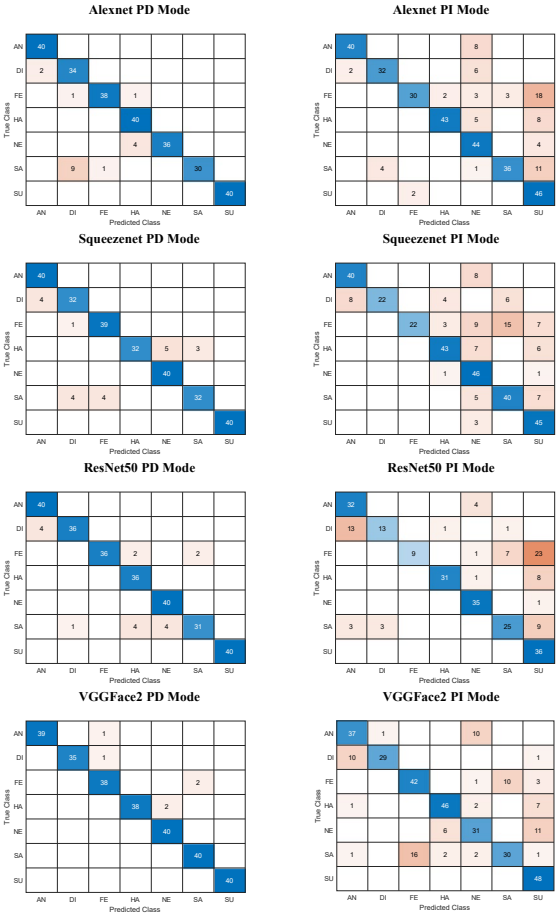

Fig. 5. Confusion matrixes for three networks in two scenarios

All simulations were repeated on the UIBVFED dataset, to evaluate five emotions, and the corresponding results for PD and PI modes are reported in Tables IV and V, respectively. Fig. 6 also provides the confusion matrixes.

TABLE IV.  RESULTS IN PD MODE FOR THE UIBVFED DATASET

| Pre-trained Net | Accuracy | Sensitivity | Specificity | Precision | F1 Score | AUC |
|---|---|---|---|---|---|---|
| AlexNet | 67.368 | 0.674 | 0.918 | 0.674 | 0.674 | 0.888 |
| SqueezeNet | 69.474 | 0.695 | 0.924 | 0.695 | 0.695 | 0.902 |
| ResNet50 | 73.684 | 0.737 | 0.934 | 0.737 | 0.737 | 0.935 |
| VGGFace2 | 71.579 | 0.716 | 0.929 | 0.716 | 0.716 | 0.929 |

TABLE V.  RESULTS IN PI MODE FOR THE UIBVFED DATASET

| Pre-trained Net | Accuracy | Sensitivity | Specificity | Precision | F1 Score | AUC |
|---|---|---|---|---|---|---|
| AlexNet | 48.936 | 0.489 | 0.872 | 0.489 | 0.489 | 0.748 |
| SqueezeNet | 52.128 | 0.521 | 0.880 | 0.521 | 0.521 | 0.737 |
| ResNet50 | 59.574 | 0.596 | 0.899 | 0.596 | 0.596 | 0.801 |
| VGGFace2 | 57.447 | 0.574 | 0.894 | 0.574 | 0.574 | 0.799 |

In this evaluation, the Resnet50 network showed the best performance. In this evaluation, the Resnet50 network showed the best performance. In PD mode, the accuracy percentage, sensitivity, specificity, precision, F1 score, and AUC were 73.68%, 0.73, 0.73, 0.93, 0.73, and 0.93, respectively. In PI mode, the accuracy percentage, sensitivity, specificity, precision, F1 score, and AUC were 59.57%, 0.59, 0.9, 0.59, 059, and 0.8, respectively. Confusion matrices provide more detailed information about the results. To our knowledge, only the article [8] mentioned the accuracy of 65% for the UIBVFED dataset without providing further details.

Based on the results, networks performed differently in various scenarios. To better understand the efficiency and shortcomings of the CNNs, the Local Interpretable Model-Agnostic Explanations (LIME) technique was used to visualize the parts of the image that the CNNs focused on when making decisions [19]. The visualization for the VGGFace2 network showed that in successful decisions, the network focused on the eyes and upper part of the face and correctly recognized the expected task.

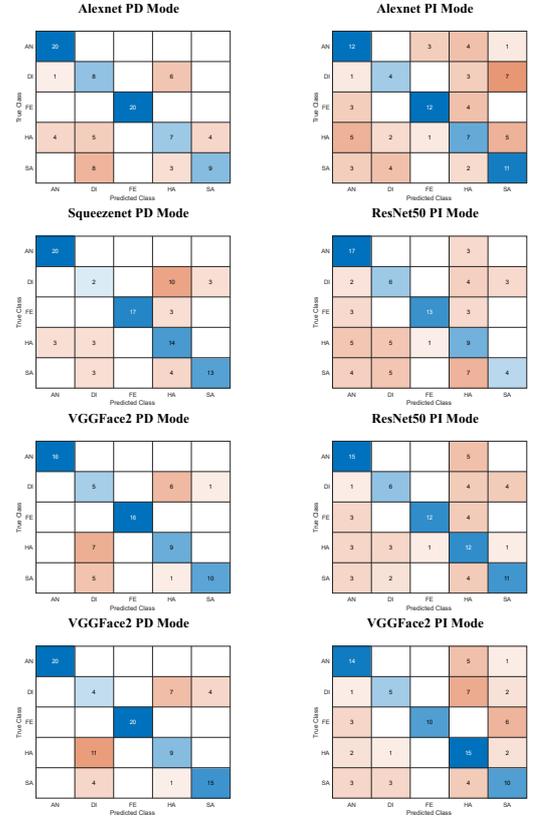

Fig. 6. Confusion matrixes for three networks in two modes for the UIBVFED dataset

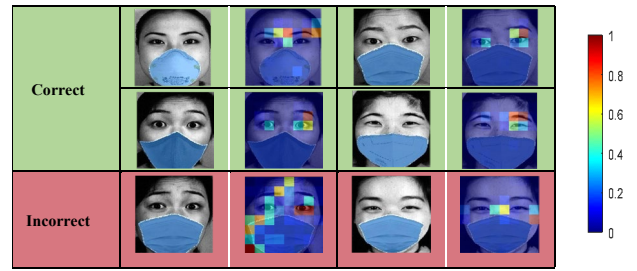

Fig. 7. Heatmap visualization of CNN feature-level using LIME for correct and incorrect decisions

However, as shown in Fig. 7, when the network focused on unimportant parts of the face, it made mistakes and was unable to recognize the emotion correctly.

## IV. CONCLUSION

In this study, we tackled the challenging task of recognizing facial emotions of individuals wearing masks using CNNs. Our approach utilized four different CNN architectures - Alexnet, Squeezenet, ResNet50 and VGGFace2 - and we evaluated our model on two datasets – JAFFE and UIBVFED. We addressed the issue of limited data availability by proposing an augmentation strategy that introduced four types of masks, increasing the diversity of images in the dataset.

Our results demonstrate that VGGFace2 provided the best performance, achieving 97.82% accuracy for PD and 74.2% for PI modes in the JAFFE dataset. For UIBVFED, the Resnet50 achieved accuracy rates of 73.68% and 59.57% for PD and PI modes, respectively. Notably, our method achieved appropriate accuracy rates while facing the additional challenge of limited data availability.

Our study contributes to the ongoing efforts in the field of AI and image processing, demonstrating the potential of CNNs to accurately recognize facial emotions even when individuals wear masks. We hope our proposed approach can serve as a baseline for further research. We have made the source code of our study available at [20], and we encourage other researchers to replicate our experiments and build upon our work.